\documentclass{article}
\usepackage{geometry}
\usepackage{graphicx}
\usepackage{subcaption}
\usepackage{booktabs}
\usepackage{amsthm,amssymb,url,amsmath}
\usepackage{mathtools}
\usepackage{ar}
\usepackage[ruled, linesnumbered]{algorithm2e}
\usepackage{color}
\graphicspath{{images/}}

\newtheorem{theorem}{Theorem}[section]

\newtheorem*{claim*}{Claim}

\newtheorem{definition}[theorem]{Definition}

\title{
Kinodynamic Planning for an Energy-Efficient Autonomous Ornithopter}

\author{Fabio Rodr\'iguez$^{2}$  \and Jos\'e-Miguel D\'iaz-B\'a\~nez$^{2,*}$ \and
      Ernesto Sanchez-Laulhe$^{1}$ \and Jes\'us Capit\'an$^{1}$
     \and An\'ibal Ollero$^{1}$
\thanks{$^{1}$GRVC Robotics Lab, University of Seville, Spain. {\tt\small jcapitan@us.es, esanchezlaulhe@us.es, aollero@us.es}}%
\thanks{$^{2}$Departamento de Matem\'atica Aplicada II, University of Seville, Spain. {\tt\small fabio25.rodriguez@gmail.com, dbanez@us.es}}%
\thanks{$*$Corresponding author}
}

\usepackage{graphicx}

\begin{document}

\maketitle

\begin{abstract}
This paper presents a novel algorithm to plan energy-efficient trajectories for autonomous ornithopters. In general, trajectory optimization is quite a relevant problem for practical applications with \emph{Unmanned Aerial Vehicles} (UAVs). Even though the problem has been well studied for fixed and rotatory-wing vehicles, there are far fewer works exploring it for flapping-wing UAVs like ornithopters. These are of interest for many applications where long flight endurance, but also hovering capabilities are required.
We propose an efficient approach to plan ornithopter trajectories that minimize energy consumption by combining gliding and flapping maneuvers. 
Our algorithm builds a tree of dynamically feasible trajectories and applies heuristic search for efficient online planning, using reference curves to guide the search and prune states. We present computational experiments to analyze and tune key parameters, as well as a comparison against a recent alternative probabilistic planning, showing best performance. Finally, we demonstrate how our algorithm can be used for planning perching maneuvers online.    
\end{abstract}


\section{Introduction}

\emph{Unmanned Aerial Vehicles} (UAVs) are spreading quite fast for many applications due to their versatility and autonomy. However, they present two main barriers to reach a wider range of applications: (i) flight endurance; and (ii) safety during interactions with people and objects in the environment. For instance, flight endurance is essential in applications like long-range inspection of infrastructures (e.g., power lines). Conventional multi-rotor UAVs do not achieve competitive flight times for those scenarios; and the use of fixed-wing UAVs does not solve the problem completely either, as capabilities for \emph{Vertical Take-Off and Landing} (VTOL) and hovering in-place are required for accurate inspections.   

Additionally, most of the aforementioned platforms are not safe enough to interact with people, due to their powerful rotor systems, their blades and the hard materials of their airframes. In order to cope with both issues, endurance and safety, bio-inspired UAVs like flapping-wing vehicles or ornithopters~\cite{grauer_acc10,nguyen_bio18,croon_ijmav09,arabagi_ijrr12} are of interest. These try to imitate birds flying, as birds can travel long distances efficiently. Thus, the main objective of GRIFFIN project~\footnote{GRIFFIN is an Advanced Grant of the European Research Council (https://griffin-erc-advanced-grant.eu).}, which is the one inspiring this work, is the design of bio-inspired flapping-wing UAVs that are able not only to fly but also to perch in order to interact with the environment through manipulation. 

A key aspect for the development of these ornithopters is to make them able to combine efficiently gliding and flapping phases, as birds do. Gliding allows the UAV to save energy and extend its flight endurance, but flapping is still necessary to increase altitude and to perform perching operations. Therefore, during the trajectory planning process, ornithopters should consider when to transition optimally between flapping and gliding, in order to save as much energy as possible. 

\begin{figure}
	\centering
	\includegraphics[width=.5\linewidth]{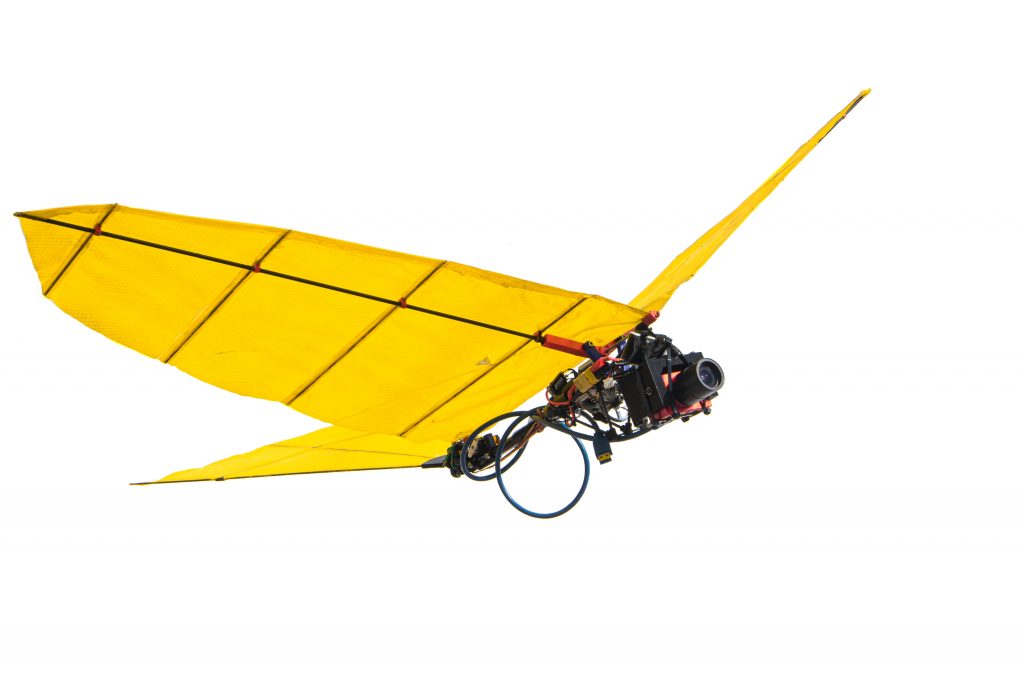}
	\caption{View of our ornithopter prototype.}
	\label{fig:prototype}
\end{figure}

In general, planning optimal trajectories for autonomous ornithopters is a complicated problem. First, these types of vehicles present nonlinear and complex dynamics that need to be taken into account when computing feasible trajectories. Second, the state space should include  vehicle's position, velocities and attitude, which are relevant for gliding and perching operations. 
Due to these complexities, there is a need for model-based but efficient methods that allow us to compute optimal trajectories with real-time performance. Some works use numerical methods for model-based trajectory planning~\cite{posa2014direct,hoff_iros19}. For instance, numerical solutions of the Navier-Stokes equations have been used~\cite{paranjape_ijas12}, but they are too expensive computationally for real-time implementation. Other approaches use probabilistic motion planners~\cite{webb_icra13,karaman_cdc10} integrating kinodynamic constraints or evolutionary algorithms~\cite{menezes_eaai16}, but again, tractable models are necessary not to exceed computational requirements.

Contrary to other related work, in this paper, we propose a novel optimization algorithm for energy-efficient trajectory planning with an autonomous ornithopter. We envision the use of flapping-wing UAVs for tasks like surveillance or inspection due to their ability to perform long-endurance flights. Therefore, our main objective is to compute online trajectories that minimize the energy consumption of the  ornithopter. Those trajectories have to comply with the ornithopter dynamics, being thus flyable. Then, we assume the existence of lower-level algorithms to control the ornithopter tracking the computed trajectory. For example, control laws for stable longitudinal and lateral flight with actual flapping-wing UAVs in perching operations have been proposed~\cite{paranjape_tro13}. 

We compute a tree of dynamically feasible trajectories by using a nonlinear model for the ornithopter motion and applying segmented tail angles and flapping frequencies.
The two flight modes of the ornithopter, i.e., flapping and gliding, are modeled with different aerodynamic coefficients, which implies a more complex nonlinear model that depends on the flight mode.
Then, we run heuristic tree search pursuing optimal solutions. In order to achieve online planning, computational complexity is alleviated twofold:
(i) we propose an ornithopter model with simplified dynamics, to make it computationally tractable; and (ii) some pruning operations to reduce the tree search space and keep it bounded. Even though we use in this paper a 2D model that constraints the ornithopter movement to a longitudinal plane, our algorithm is general and could be applied to 3D trajectory planning given a proper ornithopter model. Moreover, our heuristic solver finds approximate solutions with minimal energy consumption, but we also demonstrate the efficacy of those trajectories regarding final target state achievement.   

In summary, our main contributions are the following:

\begin{itemize}
    \item We propose a dynamic model for the 2D motion of our prototype ornithopter (see Figure~\ref{fig:prototype}). The model is nonlinear and complex, combining the aerodynamic behavior from both gliding and flapping phases. Nonetheless, we show how the model is computationally tractable for online trajectory planning. 
    \item We contribute with a new tree-based algorithm for the computation of trajectories that minimize energy consumption and comply with dynamics for the ornithopter. Gliding and flapping operations are integrated for efficient trajectory planning.
    \item Our algorithm can compute online approximated solutions by means of heuristic operations that prune the tree. We design energy-efficient curves for the ornithopter that guide the tree growth and keep the number of tree nodes bounded. 
    \item We run extensive computational experiments to analyze our algorithm and tune its key parameters. Moreover, we demonstrate its performance in comparison with another relevant approach from the literature and show a special case study for planning perching maneuvers. 
\end{itemize}

The remainder of the paper is organized as follows: Section~\ref{sec:relatedwork} surveys related work; Section~\ref{sec:problem} introduces our trajectory planning problem; Section~\ref{sec:model} describes the dynamic model for our ornithopter; Section~\ref{sec:solution} details our algorithm for trajectory planning; Section~\ref{sec:experiments} presents computational experiments to select parameters; Section~\ref{sec:results} shows experimental results to better assess the performance of our algorithm; and Section~\ref{sec:conclusions} discusses our results and explores future work. 

\section{Related Work}
\label{sec:relatedwork}

Generally speaking, motion planning for UAVs is a different problem to ours but somehow related. In the literature, the terms motion planning and path planning are usually employed indistinctly to refer to the same problem: given a robot in a workspace with obstacles, find a collision-free path from an initial to a goal configuration. 

Many methods for motion planning make use of the differential flatness of multicopter systems to generate optimal, continuous-time trajectories represented as polynomials~\cite{mellinger_icra11,mueller_tro15,oleynikova_iros16}. It can be assured that those trajectories are dynamically feasible given simplified multicopter dynamics. Others~\cite{brecianini_tro18,paranjape_ijrr15} use motion primitives to discretize the UAV state space into a connected graph. Then, standard graph search algorithms like A* can be used to find efficient solutions through the graph. 
However, the common assumptions made by the previous works do not hold for flapping-wing UAVs, where nonlinear, complex dynamics are of utter importance when planning trajectories. Therefore, instead of exploring motion planning approaches more tailored to collision avoidance, we focus on trajectory optimization methods able to take into account nonlinear dynamics in a computationally tractable manner. 

In general, the trajectory optimization problem for UAVs consists of finding the sequence of control inputs that minimizes a certain cost index, such as energy consumption or flight time; but fulfilling at the same time constraints on the vehicle dynamics. Thus, trajectory planners use the UAV motion equations (typically differential equations) to provide as output time-indexed variables such as positions, velocities and accelerations. A survey on trajectory optimization for UAVs can be found in~\cite{coutinho2018unmanned}.

A common approach for UAV trajectory planning are numerical methods. In particular, these methods can be classified into direct and indirect methods. Direct methods rely on discretizing an infinite-dimensional optimization problem into a finite-dimensional problem, to apply then nonlinear programming solvers~\cite{posa2014direct}. Indirect methods do the opposite, they first determine optimal control necessary conditions for the problem, and then use a discretization method to solve resulting equations~\cite{betts2010practical}.      
For instance, \cite{dietl_ast13} propose a discrete-time optimization method (with fixed time-step) for ornithopter trajectory optimization, where the objective is minimizing travelled distance. Another discrete numerical framework for solving constrained optimization problems using gradient-based methods is presented in~\cite{wang_aiaa17}. The technique is applied to energetically optimal flapping using frequency and pitching/heaving trajectories as optimization parameters. A longitudinal model for the dynamics of a bat-like prototype has also been proposed recently~\cite{hoff_iros19}. The authors use that model with direct collocation to plan dynamically feasible trajectories in simulation and track them with their actual prototype. The objective is reducing control efforts by minimizing accelerations.
The main concern with these numerical methods is that they can face computational and convergence issues for highly nonlinear problems, as the one addressed in this paper. This makes them less suitable for online trajectory planning. 

An alternative approach to numerical methods are probabilistic planners like \emph{Probabilistic Road-Maps} (PRM) or \emph{Rapidly-exploring Random Trees} (RRT). These algorithms are able to tackle high-dimensional planning problems in reasonable computation time by increasingly sampling the state space. Besides, they are probabilistically complete, i.e., they converge to a solution (if it exist) with probability approaching 1. There are also versions like RRT*~\cite{karaman_ijrr11} that achieve optimal solutions in the same asymptotically manner.

Algorithms based on RRT* build a tree by connecting the samples from the state space through optimal trajectories (i.e., solving the two-point boundary value problem). However, computing feasible trajectories for kinodynamic systems is an issue.
Some works have extended RRT* focusing on simple specific instances of kinodynamic systems~\cite{karaman_cdc10}. In~\cite{webb_icra13}, for example, they propose a kinodynamic RRT* that can cope with any system with controllable linear dynamics. They even apply the algorithm to nonlinear dynamics through linearization, but without convergence guarantees. 

Other probabilistic methods deal with nonlinear systems more specifically. For instance, connecting tree nodes using trajectories based on splines that are optimized via a nonlinear program solver~\cite{stoneman_cdc14}. Conversely, there exists the alternative of using exclusively control sampling to handle dynamic constraints, rather than resorting to a numerical two-point boundary value problem solver. This approach is followed by the variants \emph{Stable-Sparse RRT*}~\cite{li_ijrr16} and \emph{AO-RRT}~\cite{hauser_tro16}. Convergence is not proven for any of these RRT* modifications.
Moreover, the aforementioned probabilistic planners try to be generic solvers. Contrary to that, we propose problem-specific heuristics that allow us to guide more quickly the search of energy-efficient trajectories for an ornithopter.  







\section{Problem Description}
\label{sec:problem}

In this section, we introduce the optimization problem to solve in order to plan trajectories with an autonomous ornithopter, as well as the main assumptions we made. 

We assume that we have an autonomous ornithopter with a known model of its dynamics. Then, we are interested in planning optimal trajectories to navigate the ornithopter from an initial to a target state. This means to compute the sequence of control actions that produces a trajectory connecting the two given states that: (i) is dynamically feasible; and (ii) minimizes the total energy consumed by the ornithopter. This can be done by combining flapping and gliding maneuvers. Moreover, we assume that the UAV is flying in an open space and, hence, we do not consider collisions with obstacles. 

More specifically, let us define the \emph{states} and control \emph{maneuvers} for our problem:

\begin{definition}
(Flight state) A flight state $s=(x,z,u,w,\theta,q)$ describes an ornithopter configuration in a given instant of time, where $x$ and $z$ are the positional values in the plane $XZ$ of the Earth reference frame, $u$ and $w$ are velocity components in the body reference frame, $\theta$ is the pitch angle and $q$ is the pitch angular velocity.
\end{definition}

\begin{definition}
(Flight maneuver) A flight maneuver is a control action performed by the ornithopter during its flight at a given flight state. We consider two degrees of freedom to define flight maneuvers: tail deflection, determined by the deflection angle $\delta$ (up and down); and wing flapping, determined by the flapping frequency $f$ (including zero value for gliding). 
\end{definition}

According to the previous definitions, a trajectory consists of a sequence of interleaved flight states and flight maneuvers. Our trajectory planning problem is constrained to a 2D movement ($XZ$ plane), as we use a longitudinal motion model for the ornithopter~\footnote{Note that our trajectory planning method could be applied to 3D as long as there were a complete 3D motion model for the ornithopter.}. In particular, we define the Earth reference frame as a global frame with the $Z$ axis (pointing downwards) representing the ornithopter altitude and the $X$ axis its longitudinal motion. We also define a body reference frame attached to the ornithopter with $X_b$ pointing forward and $Z_b$ downwards. Both reference frames, together with the state and control variables are depicted in Figure~\ref{fig:ornithopterScheme}.   
Finally, we made some additional assumptions to simplify the ornithopter dynamics and derive its model: (i) we use the hypothesis of punctual mass; (ii) we assume small flapping amplitudes and thin airfoils to model aerodynamics; and (iii) we consider the aerodynamics centers to be in a fixed position, as movements are of small amplitude.

\section{Ornithopter Dynamic Model}
\label{sec:model}

We describe in this section a motion model based on the one defined in~\cite{MartinICUAS}, and specifically developed for bio-inspired, flapping-wing UAVs. The model is used to describe longitudinal motion of our ornithopter prototype.

\subsection{Non-dimensional Newton-Euler equations}

\begin{figure}
	\centering
	\includegraphics[width=.5\linewidth]{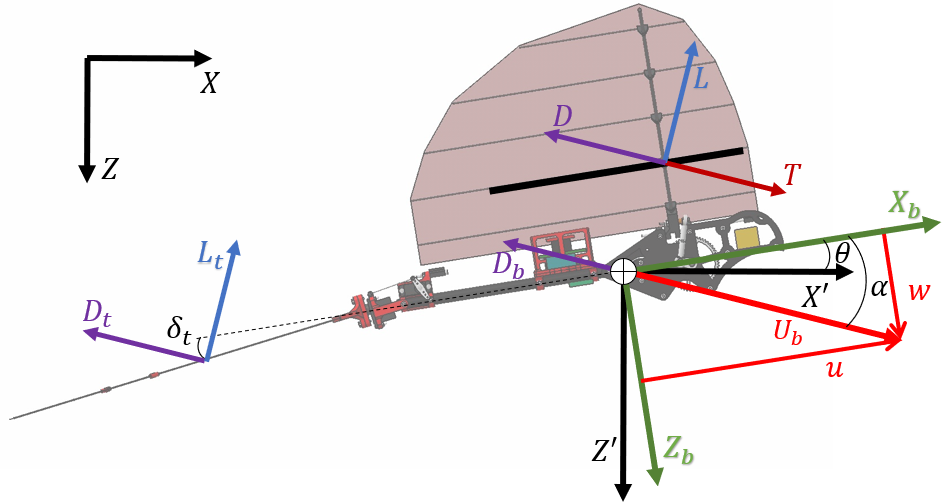}
	\caption{Schematics of the ornithopter with the forces acting on it. Axis $XZ$ represents the Earth frame; axis $X'Z'$ a translation of the Earth frame; and axis $X_bZ_b$ the body frame.}
	\label{fig:ornithopterScheme}
\end{figure}

The Newton–Euler equations describe the combined translation and rotational dynamics of a rigid body, considering all existing forces. For a flapping-wing UAV as the one in Figure~\ref{fig:ornithopterScheme}, the equations can be described as follows:

\begin{align}
    2\mathcal{M}\dfrac{du}{dt} & =U_{b}^{2}[(C_{L}+\Lambda C_{Lt})\sin{\alpha} \nonumber\\
    &+(C_{T}-C_{D}-Li-\Lambda C_{Dt})\cos{\alpha}] \nonumber \\
    &- \sin{\theta} - 2\mathcal{M}qw \label{ueq}\\
    2\mathcal{M}\dfrac{dw}{dt} & =U_{b}^{2}[-(C_{L}+\Lambda C_{Lt})\cos{\alpha} \nonumber \\
    &+(C_{T}-C_{D}-Li-\Lambda C_{Dt})\sin{\alpha}] \nonumber \\
    &+ \cos{\theta} + 2\mathcal{M}qu \label{weq}\\
    \frac{1}{\chi U_b^2}\frac{dq}{dt} & = C_L\cos(\alpha)-\left(C_{T}-C_{D}\right)\sin(\alpha) \nonumber \\
    &+\mathcal{L}\Lambda [C_{Lt}\cos(\alpha)+C_{Dt}\sin(\alpha)] \nonumber \\
    &-\mathcal{R}_{HL}[C_L\sin(\alpha)+\left(C_{T}-C_{D}\right)\cos(\alpha)] \label{qeq} \\
	\frac{d\theta}{dt} & = q, \label{thetaeq}
\end{align}

\noindent where $\alpha$ is the angle of attack, defined as $\alpha=\arctan(w/u)$, and $U_{b}=\sqrt{u^2+w^2}$ the velocity module. $\mathcal{M}$, $\chi$, $\Lambda$, $\mathcal{L}$ and $\mathcal{R}_{HL}$ are  characteristic non-dimensional parameters of the UAV. These parameters are obtained by scaling the variables with the characteristic speed, length and time:

\begin{equation}
U_c=\sqrt{\frac{2mg}{\rho S}},\qquad L_c=\frac{c}{2},\qquad t_c=\sqrt{\frac{\rho Sc^2}{8mg}},
\end{equation}

\noindent where $m$ is the mass of the UAV, $\rho$ the air density, $S$ the wing surface, $c$ the mean aerodynamic chord and $g$ the gravity acceleration. 

Figure~\ref{fig:ornithopterScheme} shows the forces acting on the vehicle, as well as the representative variables and reference frames used. In particular, $L$, $T$ and $D$ represent the lift, thrust and drag forces due to the wings; $L_t$ and $D_t$ the lift and drag from the tail; and $D_b$ the drag due to body friction. All these forces are considered in Equations \eqref{ueq}-\eqref{thetaeq} by means of the non-dimensional aerodynamic coefficients: $C_L$, $C_T$ and $C_D$ for the wing; $C_{Lt}$ and $C_{Dt}$ for the tail; and $Li$ for the body. $Li=\frac{S_b}{S}C_{Db}$ is the Lighthill's number, with $S_b$ the body surface and $C_{Db}$ its friction drag coefficient. Section~\ref{sec:aerodynamic} explains how to compute the rest of coefficients for the aerodynamic forces of the wing and the tail.

\subsection{Aerodynamic models} \label{sec:aerodynamic}

First, let us concentrate on the computation of the lift and thrust forces from the wings.
Our model considers two modes of flight for the ornithopter: flapping and gliding; and  the computation of the aerodynamic coefficients differs from one to another. In both cases, we make approximations assuming very thin airfoils. For gliding, the Prandtl's lifting line theory, combined with unsteady aerodynamic terms is used. For flapping, the Theodorsen solution~\cite{Theodorsen} defines the lift provided a wing movement with the form $h(t)=h_0\cos{(2\pi ft)}/2$, being $h$ the vertical position of the reference wing chord during the flapping movement and $h_0$ the movement amplitude. In order to model the existing thrust (only existing in flapping mode), we use the Garrick coefficient~\cite{garrick} corrected by ~\cite{FernandezF1,FernandezF2}. Moreover, we consider finite wing effects by making aspect ratio corrections based on~\cite{ayancik}, which leads to:

\begin{align}
    C_{L_{glide}}&=2\pi \left[\alpha + \left(\dfrac{1.5 \dot{\alpha}-\frac{2l_w}{c}q}{U_b}\right)\right] \frac{\AR}{\AR+2} \\
    C_{L_{flap}}&=2\pi \{\left(kh_0\right)[G(k)\cos{2\pi ft} \nonumber \\
    &+F(k)\sin{(2\pi ft)}] + \alpha\} \frac{\AR}{\AR+2} \nonumber \\
    &+ \pi k^2 h_0 \cos{(2\pi ft)}\frac{\AR}{\AR+1} \\
    C_{T_{flap}}&= 4\left(kh_0\right)^2\sin{(2\pi ft)}[F_1(k)\cos{(2\pi ft)} \nonumber \\
    &-G_1(k) \sin{(2\pi ft)} ]\frac{\AR}{\AR+2}-\alpha C_{L_{flap}} ,
\end{align}

\noindent where $\AR$ is the aspect ratio of the wing, given by $\AR=b^2/S$, being $b$ the wingspan and $S$ the surface. $l_w$ is the distance between the center of gravity and the aerodynamic center of the wing, being positive when the center of gravity is behind the wing.
$k=2\pi f/U_b$ is the reduced frequency; $F(k)$ and $G(k)$ are the real and imaginary parts of the Theodorsen's function $C(k)$; and $F_1(k)$ and $G_1(k)$ are the real and imaginary parts of the function $C_1(k)$ defined in~\cite{FernandezF1}. 

Regarding the lift force generated by the tail, as our bio-inspired design leads to triangular surfaces~\cite{thomastail}, we use an approximation for delta wings:

\begin{equation}
    C_{L_t} \hspace{-1pt}= \hspace{-1pt}\dfrac{\pi \AR_t}{2} \hspace{-2pt} \left[(1 \hspace{-1pt}-\hspace{-1pt} \varepsilon_{\alpha})\alpha \hspace{-1pt}+\hspace{-1pt} \delta \hspace{-1pt}+\hspace{-1pt} \left(\dfrac{1.5 \dot{\alpha}-\frac{2l_t}{c}q}{U_b}\right)\right] ,
\end{equation}

\noindent where $\delta$ is the deflection angle of the tail; $\varepsilon_{\alpha}$ models the interference caused by the wing; and $\AR_t$ and $l_t$ are, respectively, the aspect ratio of the tail and the distance between the center of gravity and the aerodynamic center of the wing, being defined in the same manner as it was for the wing.
In order to consider near stall effects, we saturate all lift coefficients ($C_L$ and $C_{L_t}$) for angles greater than $10^o$ for the wing and $25^o$ for the tail.

Finally, we model the drags from wings and tail, $C_{D}$ and $C_{Dt}$, as the addition of constant friction drags, $C_{D0}$ and $C_{D0t}$, and induced drags, provided by:

\begin{equation}
    C_{D_i}=\frac{C_{L}^2}{\pi\AR},\qquad C_{D_{it}}=\frac{C_{L_t}^2}{\pi\AR_t}.
\end{equation}

\section{Ornithopter Segmentation-based Planning Approach}
\label{sec:solution}

In this section, we present our method to solve the problem stated in Section~\ref{sec:problem}, i.e., to plan optimal trajectories for an ornithopter that are both dynamically feasible and energy-efficient.
As discussed in Section~\ref{sec:relatedwork}, there exist in the literature numerical optimization solvers that can deal with nonlinear systems. They obtain dynamically feasible trajectories by discretizing and integrating the model dynamics, but can suffer from computational complexity and convergence issues for highly nonlinear models. Graph-based approaches are an alternative, where a graph with discrete, connected states is built in order to search for optimal trajectories. In particular, probabilistic planners~\cite{karaman_ijrr11} sample the state space increasingly to build these graphs. If the objective is to generate trajectories that are dynamically feasible, the method must ensure that the sampled states are reachable, which can be complex for nonlinear systems.

In this paper, we propose a novel graph-based approach that builds a tree to search for energy-efficient trajectories. Instead of taking random samples from the state space, we segment the ornithopter's actions and integrate its dynamics to generate and connect feasible states, which can later become nodes of the tree. 
We named our algorithm \emph{OSPA}, which stands for \emph{Ornithopter Segmentation-based Planning Approach}.

\begin{figure}
  \centering
  \begin{subfigure}[b]{.5\textwidth}
    \includegraphics[width=8cm]{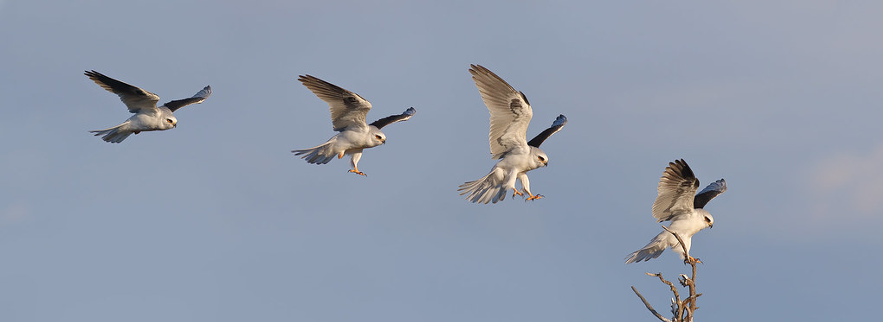}
    \label{fig:discre}
  \end{subfigure}
  \begin{subfigure}[b]{.5\textwidth}
    \includegraphics[width=8cm]{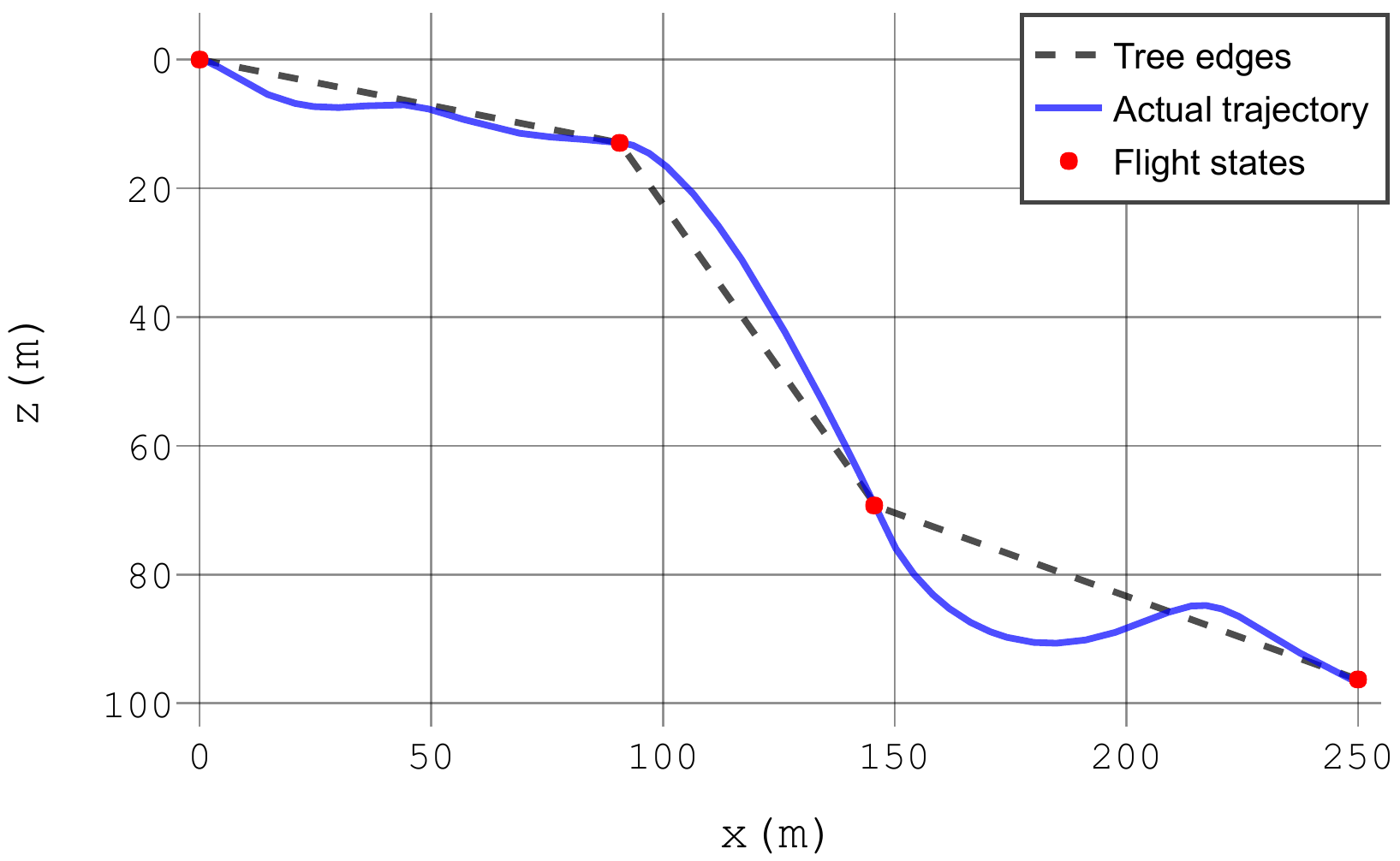}
    \label{fig:trajectory}
  \end{subfigure}
  \caption{Segmentation of the ornithopter motion for a landing trajectory. Top view, sequence of flight states followed by a real bird. Bottom view, trajectory computed by our method, with three consecutive maneuvers for 250 meters before landing. The trajectory connects the flight states by integrating the dynamic model.}
  \label{fig:segmentation}
\end{figure}

The general idea is as follows. In our approach, we consider a discrete set $M$ with the possible flight maneuvers for the ornithopter. Then, given a fixed time step $t_s$ and an initial state, we generate a discrete set of reachable flight states by integrating the ornithopter dynamics for time $t_s$ and for each maneuver in $M$. Doing that iteratively, we build a tree $T$ whose vertices or nodes are flight states and each edge has associated the corresponding maneuver to navigate from one state (node) to the next one. Moreover, each edge has also associated the energy consumed by the corresponding maneuver to transition between its vertices. We store at each vertex state the energy consumption to reach it from the initial state. The final goal is to find a path $\tau$ through the tree $T$ that connects the initial and target states and minimizes the total energy required by the ornithopter.
Figure~\ref{fig:segmentation} illustrates an example on how to segment the trajectory of an ornithopter and the resulting flight states. A landing operation is achieved by a maneuver with flapping involved, followed by two different maneuvers where the ornithopter is only gliding.   

Algorithm~$\ref{alg:path-planning}$ provides an overview of the procedure followed by OSPA.
The algorithm receives as input: \begin{itemize}
    \item the initial and final states $s_0$ and $s_f$, respectively; 
    \item the discrete set of maneuvers $M$;
    \item the time step $t_s$; 
    \item pruning parameters $k_d$ and $k_w$.
\end{itemize}
The set $M$ is generated by combining a discrete set of flapping frequencies with a discrete set of tail angles, i.e., $M=D\times F = \lbrace (\delta,f)\; | \; \delta \in D \wedge f \in F \rbrace$. 
The initial state is inserted as the tree's root. Then, at each iteration, all leaf states are expanded with all possible maneuvers. Given a state $s_i$, a flight maneuver defined by tail angle $\delta$ and frequency $f$, and a time step $t_s$, the function \texttt{ODE($s_i, \delta, f, t_s$)} integrates the equations in Section~\ref{sec:model} for time $t_s$ to produce a flight state that becomes a new node of the tree. The states that are eventually inserted into the tree are decided according to two pruning procedures that will be detailed in Section~\ref{sec:treeReduction}.

\begin{algorithm}[h]
\DontPrintSemicolon
\SetAlgoLined
\SetKwInOut{Input}{Input}
\SetKwInOut{Output}{Output}   
\Input{$(s_0, s_f, M, t_s, k_d, k_w)$}
\Output{$\tau^*$}
$tree \gets \texttt{Tree()}$\;
$tree.root \gets s_0$\;
$leaves \gets \texttt{GetLeaves($tree$)}$\;
$corridor \gets \texttt{GetCorridor($s_0$,$s_f$)}$\;
\While{$leaves.length > 0$}{
$states \gets \texttt{List()}$\;
\For{$s_i$ in $leaves$}{
\For{$(\delta,f)$ in $M$}{
$s' \gets \ \texttt{ODE(}s_i, \delta, f, t_s\texttt{)}$\;
\uIf{$ \mathtt{GetDist}(corridor, s') \leq k_d$ $and$ $s'.x \leq s_f.x$}{
$s'.parent \gets s_i$\;
$states.Add(s')$\;}
}}
$partitions \gets \texttt{GetPartitions($states$,$k_w$)}$\;
\For{$c$ in $partitions$}{
$s^* \gets \texttt{GetOptimalState($c$)}$\;
$s_i \gets s^*.parent$\;
$s_i.AddChild(s^*)$\;}
$leaves \gets \texttt{GetLeaves($tree$)}$\;
}
$\tau^*$ $\gets$ \texttt{SearchOptimalPath($tree, s_f$)}\;
\KwRet $\tau^*$
    \caption{OSPA}
    \label{alg:path-planning}
\end{algorithm}

We assume that the ornithopter has forward motion in the $XZ$ plane, so the method generates states with increasing $X$-axis values between time steps. The tree computation ends when all reached states have greater $X$-axis value than the final state $s_f$. In that case, the best sequence of maneuvers is returned and the algorithm terminates. This is done by the function \texttt{SearchOptimalPath()}, that computes the tree path $\tau^*$ with minimum energy consumption. The last node of the solution $\tau^*$ is chosen as the one with lowest energy consumption among those within a \emph{tolerance} distance to $s_f$ that is set by the user.

OSPA searches for optimal trajectories in terms of energy, and hence, whenever a new node is added, the algorithm computes and stores the accumulated energy at that node. In order to model the energy consumed by the ornithopter performing a certain maneuver for a time step $t_s$, we use the following formula: 

\begin{equation}
E = t_s(K_{aero}f^3 + c_r).
\label{eq:energy}
\end{equation}

The first term represents the dominant energy consumption, which is due to flapping wings. Using equations in Section~\ref{sec:model} with some simplifications~\footnote{Due to space limitations, we do not reproduce here the tedious proof.}, it can be proven that this consumption is proportional to the cube of the flapping frequency, with a constant coefficient $K_{aero}$ that depends on several physical characteristics of the ornithopter, such as the wings' profile, their inertia and their movement amplitude. However, as modelling all those effects precisely is complicated, we opted for estimating the value of $K_{aero}$ empirically~\footnote{All the experiments in this paper used a value $K_{aero} = 2.5 W / Hz^3$, obtained empirically for our ornithopter prototype.}. The second term models the residual energy consumption $c_r$ when the ornithopter is not flapping, mainly due to the onboard electronics. During gliding, we measured empirically for our ornithopter that the cost of moving the tail was negligible compared to the average electronics consumption. Therefore, we consider the cost $c_r$ constant~\footnote{We estimated empirically an upper bound of $c_r=5W$.}. As expected, it is important to note that Equation~\ref{eq:energy} indicates that the main energy consumption is produced by flapping maneuvers. For gliding, the energy efficiency is related with the temporal length of the maneuver. 

Let us now analyze the size of the tree $T$ generated by our method. If we have $|M| = |D|\times|F|=m$ different maneuvers that can be selected at each iteration, the whole tree construction without pruning operations takes $O(h^{rm})$ time, where
$h$ is the tree height~\footnote{The height of $T$ is the maximum distance (number of edges) from the root to any node in $T$.} and $r$ is the average time needed by the integrator \texttt{ODE()}. Note that the height $h$ depends on the time step $t_s$; the smaller $t_s$, the larger the tree will be to reach the final state. Our OSPA algorithm can achieve energy-efficient trajectories if we take values of the time step short enough and use enough number of maneuvers. However, the computational complexity increases exponentially with the number of maneuvers, and a more reduced set of states may be enough to achieve competitive approximate solutions to the final state. 
Therefore, in the next section we propose two pruning procedures which alleviate the computational cost of the algorithm and yield an efficient planner for short and medium distance flights.

\subsection{Tree reduction}
\label{sec:treeReduction}
\begin{figure}[h]
	\centering
	\begin{subfigure}{0.48\textwidth}
		\includegraphics[width = 8cm, height=6cm]{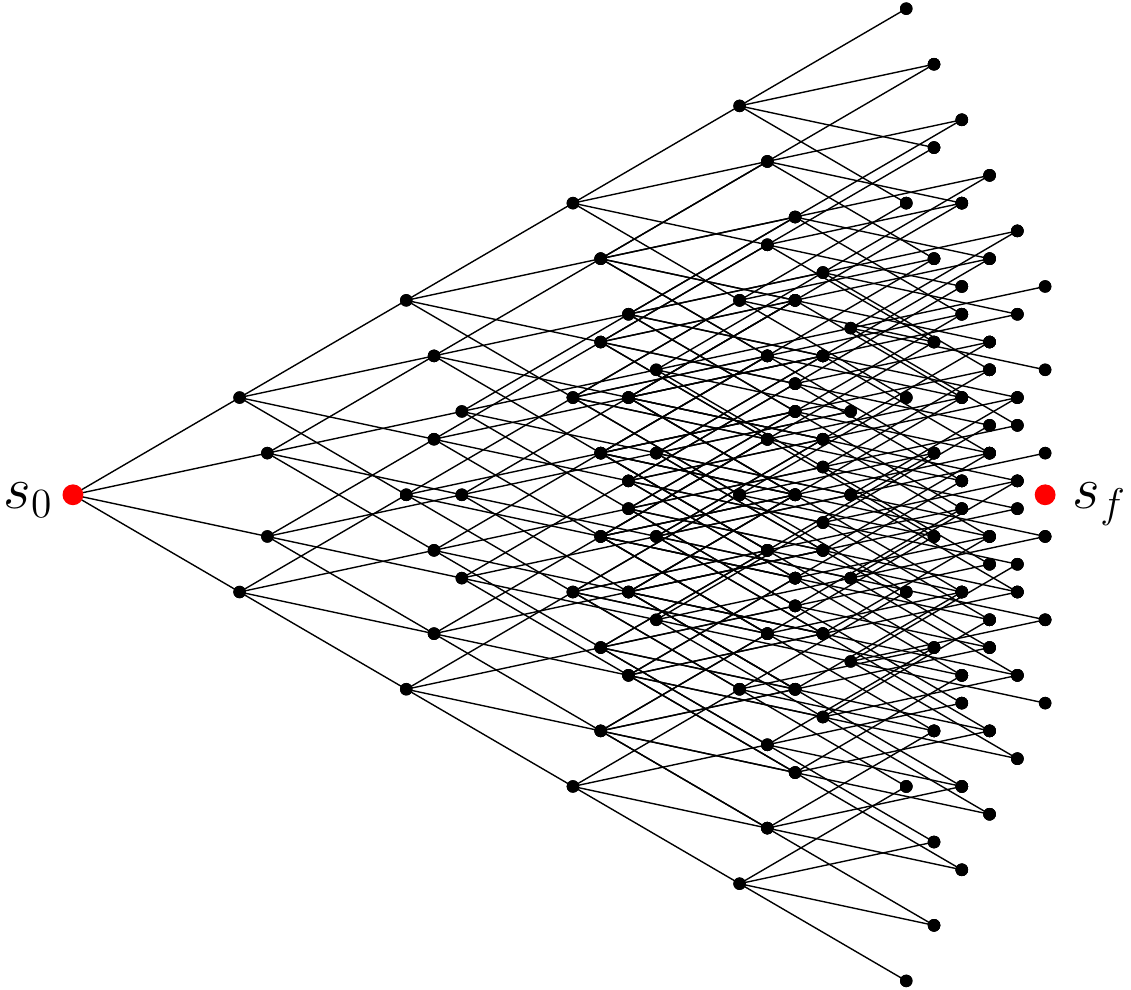} 
		\caption{Example of an exhaustive tree $T$ considering 4 maneuvers. Nodes are represented as positional values in the $XZ$ plane.}
		\label{fig:treeImg1}
	\end{subfigure}
	
	\begin{subfigure}{0.48\textwidth}
		\includegraphics[scale=0.7]{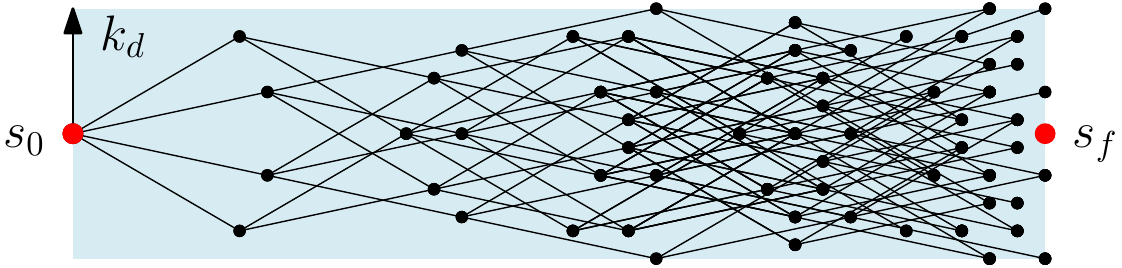}
		\caption{Tree $T'$ after applying to $T$ the pruning operation based on corridor. $k_d$ is the width or clearance of the corridor.}
		\label{fig:treeImg2}
	\end{subfigure}
	
	\begin{subfigure}{0.48\textwidth}
		\includegraphics[scale=0.7]{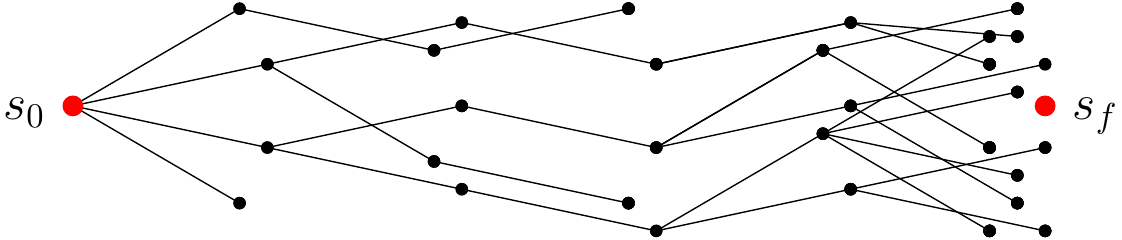}
		\caption{Tree $T''$ after applying to $T'$ the pruning operation based on witnesses. $k_w=5$ is the upper bound on the number of considered partitions at each step (the last layer of leaves is not pruned).}
		\label{fig:treeImg3}
	\end{subfigure}
	
	\caption{Overview of the pruning operations in our OSPA algorithm.}
	\label{fig:wholetree}
\end{figure}

OSPA includes two specific procedures to reduce the tree size and speed up the algorithm. First, we reduce the original tree $T$ to a pruned tree $T'$ that only keeps nodes whose position is close to a hypothetical optimal trajectory $\mathcal{P}^*$. This pruning operation relies on the idea that the tree $T'$ will produce optimal solutions similar to those in $T$, as long as the pruned nodes are not in the vicinity of $\mathcal{P}^*$. In principle, this optimal trajectory is unknown, but we propose a parametric curve to estimate $\mathcal{P}^*$ that acts as guide in the tree pruning. Second, considering many states that are very close in the tree may lead to redundant calculation of similar trajectories. This fact motivates our second procedure to reduce further the tree size, where we obtain a new tree $T''$ by creating partitions with the nodes in $T'$ and keeping only a witness node for each partition. Note that our pruning procedures assume that there is an optimal trajectory $\mathcal{P}^*$ which has robust clearance, that is, the nodes in $T''$ are enough to compute a near optimal trajectory. We support this assumption experimentally in Section~\ref{sec:experiments}. 

Figure~\ref{fig:wholetree} shows an example of an original tree $T$ and the effect of applying our pruning operations to obtain the trees $T'$ and $T''$. Note that, some of the leaves in $T''$ are still irrelevant flight states for a suitable solution since they are far away from the final state. This is why OSPA only considers for last state selection those within a desired tolerance distance to the target.
In the following, we elaborate on our two pruning operations to perform the incremental construction of $T''$.

\subsubsection{First pruning operation: the corridor}

The first procedure to prune the tree consists of imposing physical constraints on the admissible trajectories. Thus, we speed up operations in OSPA but also avoid pathological states that would be unlikely. More precisely, we define a corridor region $C$ connecting the initial and final states and only consider tree nodes within that corridor, discarding those out of the corridor. The key idea is building $C$ in such a manner that the (unknown) optimal trajectory $\mathcal{P}^*$ is likely to lie within that corridor.

We generate the corridor $C$ as follows. First, we take a parametric curve on the $XZ$ plane that connects the initial and final states, and adjust its parameters so that the curve is likely to resemble the optimal trajectory $\mathcal{P}^*$. Let this \emph{reference curve} be denoted as $\widehat{\mathcal{P}^*}$, then $C$ is defined as the region of points in the $XZ$ plane whose minimum Euclidean distance to $\widehat{\mathcal{P}^*}$ is not greater than $k_d$. Thus, the tree $T'$ results from pruning all nodes in the original $T$ that are out of $C$. Figure~\ref{fig:treeImg2} illustrates an example for a line segment $\widehat{\mathcal{P}^*}$ connecting the initial and final states. Moreover, the functions \texttt{GetCorridor()} and \texttt{GetDist()} in Algorithm~$\ref{alg:path-planning}$ compute the corridor $C$ and the minimum Euclidean distance of a flight state to $\widehat{\mathcal{P}^*}$, respectively.

Recall that the actual optimal $\mathcal{P}^*$ is unknown, but we need a fairly reasonable estimation $\widehat{\mathcal{P}^*}$ so that OSPA finds solutions that are close to the optimum. An option to learn those curves would be a totally bio-inspired approach, i.e., observing bird flights in order to copy the type of curves they follow; as they are assumed to be energy-efficient. This would require gathering large datasets which are not easy to obtain in general, so we followed a different empirical strategy to design our reference curves. Particularly, we ran extensive computational experiments in different situations, using our algorithm OSPA without pruning to compute the trajectories with lowest energy consumption. Then, we observed that trigonometric curves were well suited to these optimal trajectories. Therefore, we build our reference curve $\widehat{\mathcal{P}^*}$ as a special type of trigonometric curve that connects the initial and final states, and assume that it is a fairly good approximation of the actual optimal trajectory. The procedure to compute and adjust these reference curves is further detailed in Section~\ref{sec:referenceCurve}.  

\subsubsection{Second pruning operation: witness states}

Our second procedure to prune the tree focuses on preventing redundancy in order to improve further the time complexity, but without degrading significantly the solution quality. The idea is the following. Note that at each tree expansion step, neighboring nodes generate $M$ new flight states each, and some of them may be similar. Indeed, the density of close states will grow as the tree height increases (see example in Figure~\ref{fig:treeImg1}). Therefore, we implement a simple partitioning technique to group close nodes and select only the best ones. At each iteration of the tree building process, we take all new leaf nodes generated and partition them into $k_w$ disjoint subsets.
Since nodes at the same tree level present more significant differences in the vertical coordinate than in the horizontal one, the $z$ coordinate is used to order leaves and split them into $k_w$ equally separated sets. This is done by function \texttt{GetPartitions()} in Algorithm~\ref{alg:path-planning}. 
As all nodes in each partition will represent close flight states, we keep alive only a \emph{witness} node for each partition, throwing away the rest. Since we search for energy efficiency, the selected witness nodes are those with minimum accumulated energy (function \texttt{GetOptimalState()} in Algorithm~\ref{alg:path-planning}). Figure~\ref{fig:treeImg3} illustrates an example of this pruning operation. 

Finally, note that this witness pruning operation alleviates considerably the computational complexity of OSPA, as the number of leaf nodes inserted at each iteration of Algorithm~$\ref{alg:path-planning}$ is bounded. More specifically, the original computation time to build the tree $O(h^{r m})$ is reduced to $O(h k_w m(r+\log{(k_w  m)})$, which is almost linear in the number of maneuvers. The second term in the time complexity is due to the leaves ordering in $z$ performed by \texttt{GetPartitions()}.


\section{Selection of Parameters}
\label{sec:experiments}

In this section, we describe a series of computational experiments to analyze the effects of the key parameters in OSPA. We aim to obtain the most appropriate values for: the time step $t_s$, the reference curve $\widehat{\mathcal{P}^*}$, the set of maneuvers $M$, the threshold $k_d$ to compute the corridor $C$ and the parameter $k_w$ to create partitions. 
All experiments were run with a version of OSPA coded in Python 3.7~\footnote{The code is available at \url{https://github.com/fragnarxx/kinodynamic-planning}.}, on a CPU with a $1.60~GHz$ processor and $8~GB$ RAM. We used the \texttt{odeint} module from the \emph{SciPy} library for numerical integration.

\begin{table}
 \centering
 \begin{tabular}{|c | l |} 
 \hline
 \textbf{Parameter}  &  \textbf{Interval} \\ [0.5ex] 
 \hline
 \hline
 $\delta$ & $[-6,0]^\circ$ \\
 \hline
 $f$ & $[0,6]~Hz$ \\
 \hline
 $t_s$ & $[8,20]~s$ \\
 \hline
 $k_d$ & $[10,25]~m$ \\ 
 \hline
 $k_w$ & $[10,40]$ \\ 
 \hline
\end{tabular}
\caption{Intervals used for the parameter values in the tuning experiments.}
\label{tab:parameters}
\end{table}

Table~\ref{tab:parameters} shows the parameter intervals used for all experiments. 
We consider a discrete set $M$ of maneuvers that results from the combination of 7 tail angles uniformly selected between the bounds in Table~\ref{tab:parameters} and the frequency values $\lbrace 0,4,5,6\rbrace~Hz$, hence $|M|=28$. These frequency values were selected to provide positive net thrust and consider the mechanical engine limitations of our ornithopter prototype. The interval for tail deflections was selected to ensure that the ornithopter flies without reaching aerodynamic stall, which would not be interesting in general flight conditions.

\begin{table}[htbp]
	\begin{center}
		\begin{tabular}{|c|c|c|c|c|}
			\hline
			$\mathcal{M}$ & $\Lambda$ & $\mathcal{L}$ & $ \mathcal{R}_{HL} $ & $ \chi $ \\
			\hline
			$6.85$ & $0.278$ & $-15.5$ & $ 1.92 $ & $ 0.0132 $ \\
			\hline
			\hline
			$C_{D_{0}}$ & $C_{D_{0t}}$ & $\AR$ & $ \AR_t $ & $Li$ \\
			\hline
			$0.018$ & $0.021$ & $4.44$ & $ 2.35 $ & $ 0.0051 $ \\
			\hline
		\end{tabular}
	\end{center}
	\caption{Values for the dimensionless characteristic parameters of the ornithopter.}
	\label{tab:ParametersOrnithopter}
\end{table}

\begin{table}[htbp]
	\begin{center}
		\begin{tabular}{|c|c|c|}
			\hline
			$U_c$ & $L_c$ & $t_c$ \\
			\hline
			$4.26~m/s$ & $0.135~m$ & $0.0317~s$ \\
			\hline
		\end{tabular}
	\end{center}
	\caption{Values for the characteristic dimensions.}
	\label{tab:dimensions}
\end{table}

In all experiments, we used the physical properties of our actual prototype in Figure~\ref{fig:prototype} to determine the parameters of the dynamic model from Section~\ref{sec:model}.
Table~\ref{tab:ParametersOrnithopter} depicts the characteristic dimensionless parameters, while Table~\ref{tab:dimensions} depicts the characteristic dimensions of the problem.
Finally, we built the same set of $80$ scenarios for all experiments, with the ornithopter starting at $(0,0)$ position in the $XZ$ plane, and $0^\circ$ for the initial and final pitch angles. The final state positions are taken from a uniform grid within the rectangle $R=\{(x,z): 200\leq x\leq 250, -20\leq z \leq 100 \}$. Recall that positive and negative values for $z$ mean descending and ascending, respectively.

\subsection{Time step}

\begin{figure}
	\centering
	\includegraphics[width=0.5\columnwidth]{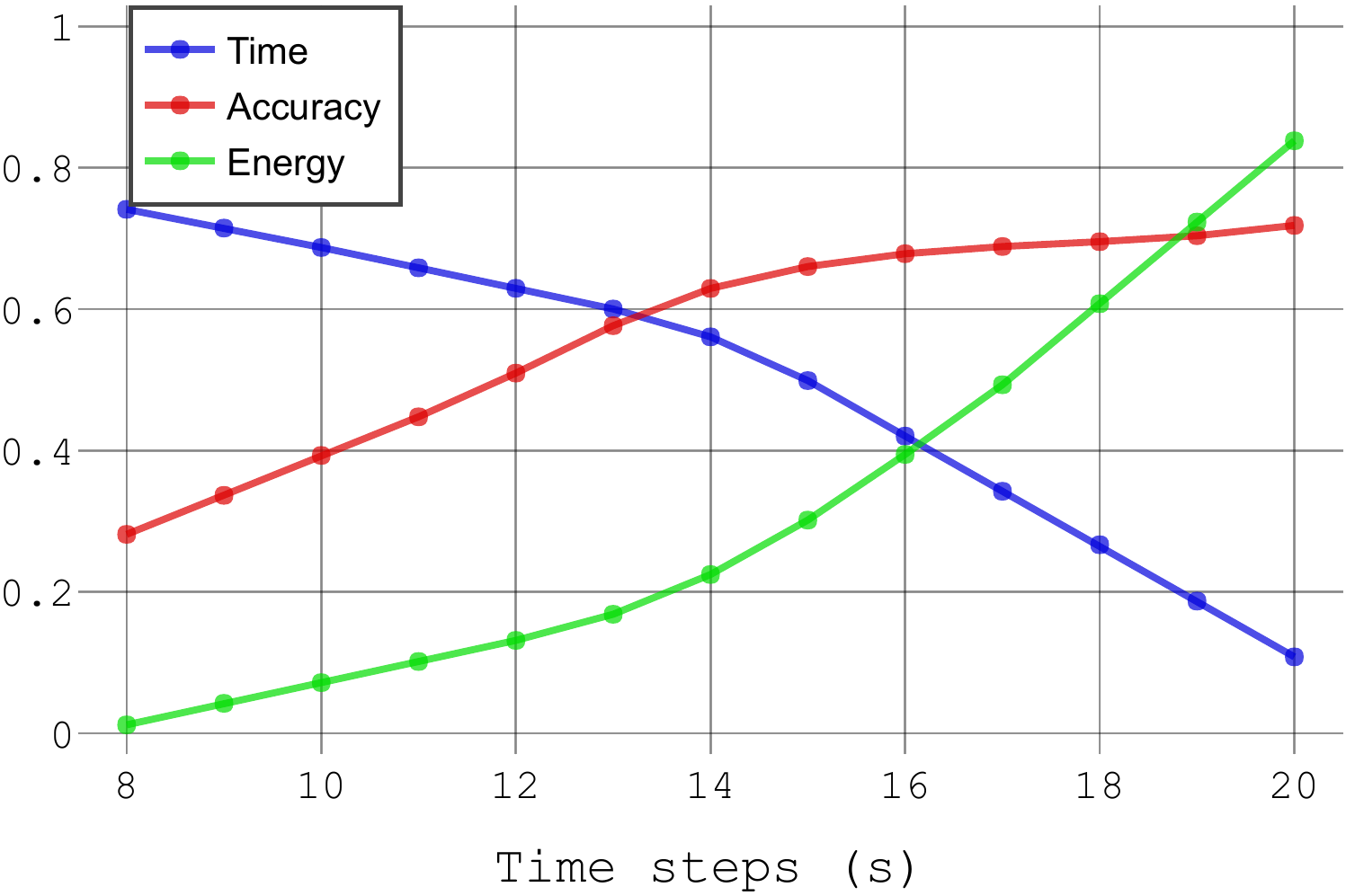}
	\caption{Performance results varying the time step $t_s$. Average values over 80 experiments are shown for the computation time, the accuracy w.r.t. the target state ($\Delta$) and the solution energy ($E$). Values are scaled to the $[0,1]$ interval.}
	\label{fig:timestep}
\end{figure}

The first critical parameter in OSPA is the time step $t_s$, as it affects the computation time, as well as the quality of the solution. In order to select an adequate value for the parameter, we ran OSPA at the $80$ experimental instances using the set of maneuvers $M$ and varying the value of $t_s$. 
Figure~\ref{fig:timestep} shows average results as $t_s$ increases. In particular, we consider three metrics: the computation time to plan a trajectory, the total energy $E$ consumed throughout the trajectory and the \emph{accuracy} $\Delta$, defined as the Euclidean distance between the last node of the trajectory and the target (recall that the nodes are vectors including positions, angles and velocities). As expected, smaller values of $t_s$ yield a larger computation time, but a higher accuracy, decreasing the error w.r.t. the target. According to this trend, we selected a value of $t_s=12~s$ as a trade-off solution, since it favors energy consumption but at the same time produces acceptable accuracy values.

\subsection{Reference curve}
\label{sec:referenceCurve}
\begin{figure}
	\centering
	\begin{subfigure}[b]{0.5\textwidth}
		\includegraphics[width=7.8cm, height=4.5cm]{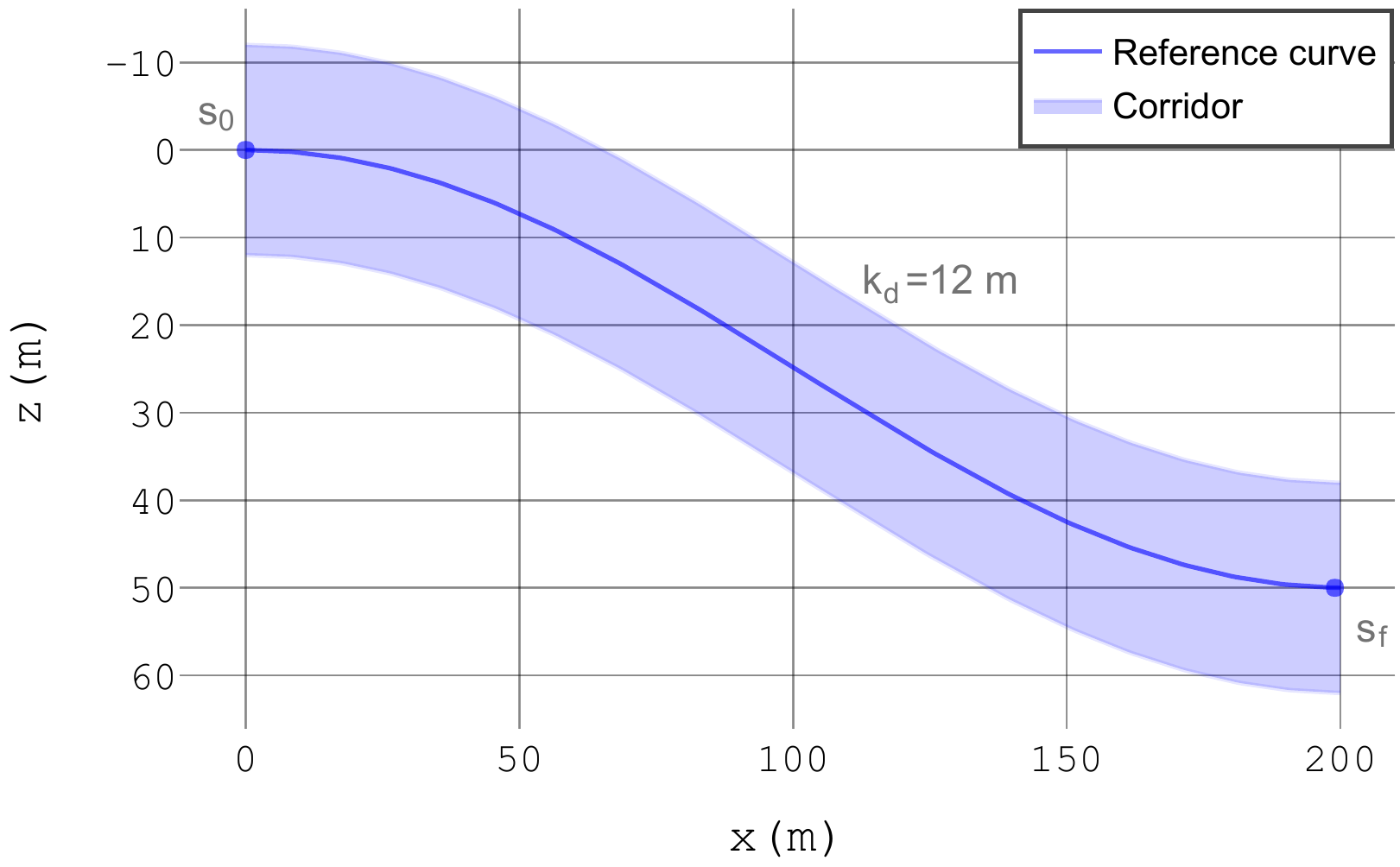}
		\label{fig:curve1} 
	\end{subfigure}
	\begin{subfigure}[b]{0.5\textwidth}
		\includegraphics[width=7.8cm, height=4.5cm]{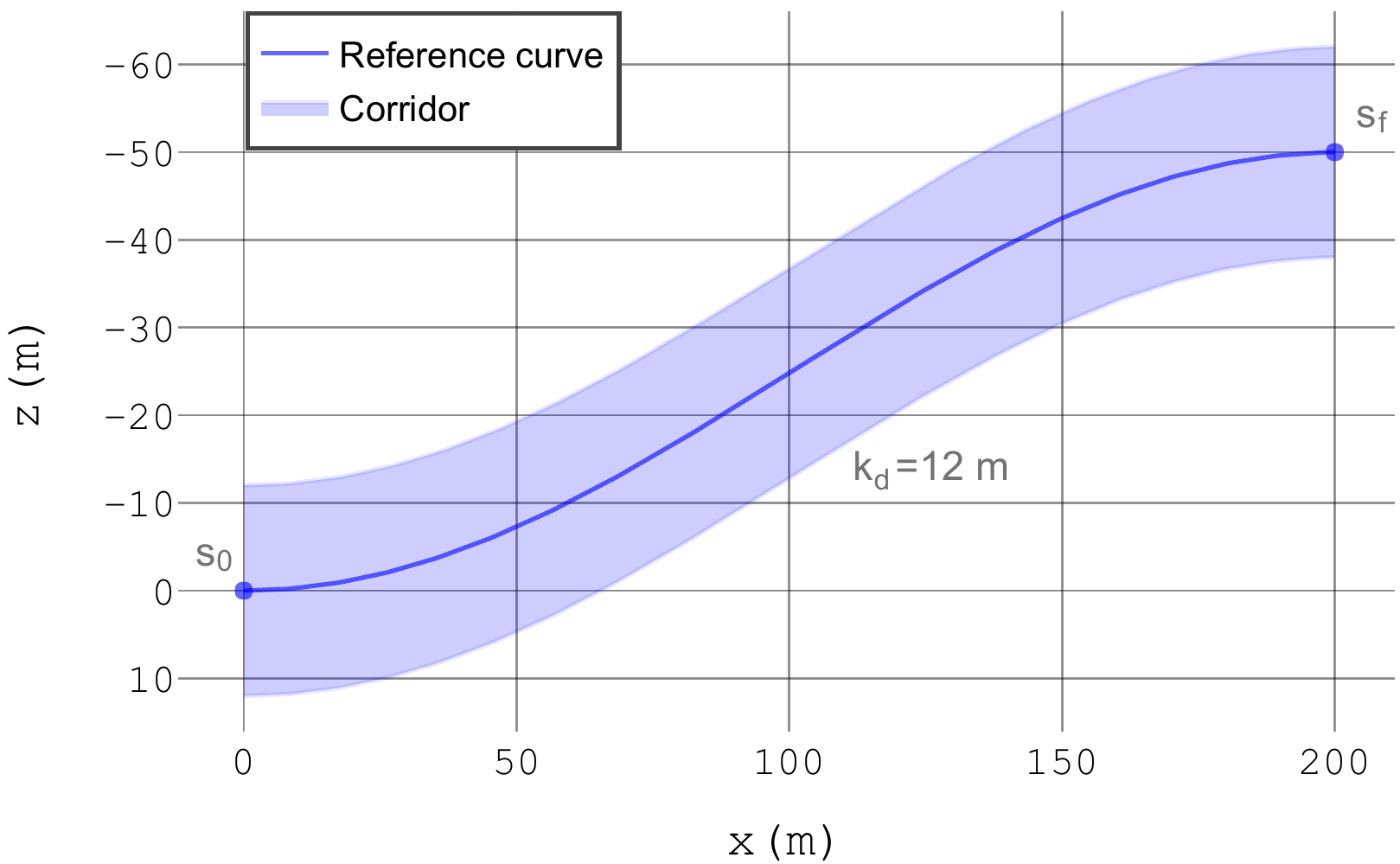}
		\label{fig:curve2}
	\end{subfigure}
	\caption{Examples of the reference curve for two cases with the ornithopter descending (top) and ascending (bottom). The corresponding corridor $C$ is also shown for $k_d=12$.}
	\label{fig:reference}
\end{figure}

In this section we describe the empirical process to design the reference curve $\widehat{\mathcal{P}^*}$ in OSPA.
Recall that this curve is used to define a corridor $C$ that guides the tree search. As we prune all states with distance greater than $k_d$ from the reference curve, we should select a curve that is as close as possible to the typical waypoints in optimal trajectories. For that, we ran the 80 experiments obtaining optimal trajectories using OSPA with $t_s=12~s$ and the set of maneuvers $M$, without pruning operations.

As a first option, we attempted a simple straight line connecting the origin and the target point. Then, we measured the \emph{Maximum Error} (ME) and the \emph{Root Squared Mean Error} (RSME) between the waypoints obtained in the trajectories computed by OSPA and the straight line, obtaining the average values $15.40~m$ and $9.24~m$, respectively.
However, we observed from our results that trigonometric curves were well suited to the trajectories generated in the experiments. Therefore, we built our reference curve $\widehat{\mathcal{P}^*}$ as a scaled cosine connecting the initial and final states. The formula for this curve depends on the distance between the initial $(x_0,z_0)$ and final $(x_f,z_f)$ positions. Provided that $x_d=x_f-x_0$ and $z_d=z_f-z_0$, the formula for the proposed reference curve is:

\begin{equation}
\widehat{\mathcal{P}^*}(x) = \frac{1}{2}\left [ z_d + z_d cos \left (\pi + \frac{\pi x}{|x_d|} \right ) \right ].   
\label{eq:refCurve}
\end{equation}

Figure~\ref{fig:reference} depicts examples of this reference curve and the corridors, resembling typical optimal trajectories obtained with our ornithopter model. Moreover, we computed the ME and RMSE metrics for our 80 experiments, obtaining the average values $15.30~m$ and $8.92~m$, respectively. As expected, we determine that our trigonometric choice of reference curve is also closer to optimal trajectories than straight lines, and we use it in the remaining experiments.

\subsection{Study of maneuvers}\label{section:maneuvers}

We defined already a set of maneuvers ($|M|=28$) that make sense from the physical point of view of our ornithopter to plan trajectories. However, as it was explained in Section~\ref{sec:solution}, the number of maneuvers affects highly the computation time of OSPA. Therefore, we perform in this section a statistical analysis to find out which maneuvers are really more useful, with the purpose of further reducing the final set. For this study, according to our previous time step analysis, we fixed $t_s=12~s$. Then, we ran OSPA (without pruning operations) at the $80$ scenarios to find optimal trajectories. We can compute the occurrence rate of a maneuver as follows:   

\begin{equation}
\xi(m_i)=n_i/|n^*|,
\end{equation}

\noindent where $n_i$ is the number of times that maneuver $m_i$ was selected in any of the optimal solutions and $n^*$ is the total count of maneuvers in all optimal solutions. Larger values of $\xi$ indicate that the maneuver is commonly used, while lower values correspond to rarely used maneuvers. Therefore, we select our reduced set of maneuvers $M_r$ by defining a threshold value $\xi^*$ and removing maneuvers with lower occurrence rate, i.e., $M_r = \lbrace m_i: m_i \in M,\: \xi(m_i) \ge \xi^* \rbrace$.


\begin{table}
 \centering
 \begin{tabular}{| c | c | c | c | } 
 \hline 
 $\xi^*$    & $|M_r|$ & $\Delta$    & $E~(W)$ \\ [0.5ex]
 \hline \hline
 $0$      & $28=|M|$  & $0.49\pm 0.06$ & $3979\pm 343$\\
 \hline
 $0.01$   & $23$ & $0.51\pm 0.07$ & $3693\pm 298$\\
 \hline
 $0.02$   & $17$ & $0.53\pm 0.07$ & $3393\pm 317$ \\
 \hline
 $0.03$   & $10$ & $1.31\pm 0.08$ & $2268\pm 254$ \\
 \hline
\end{tabular}
\caption{Results with different sets of maneuvers. Average values and deviations over 80 simulations are shown for the accuracy ($\Delta$) and the energy ($E$).}
\label{tab:thresholds}
\end{table}

We tested different values of $\xi^*$ to create the subset $M_r$, and solved again the 80 scenarios with each $M_r$. Table~\ref{tab:thresholds} depicts average results for the accuracy and energy as $\xi^*$ increases.
The computation time is not included because it is not representative (the experiment uses the whole set of maneuvers $M$ without pruning operations, and hence, it is not efficient). According to the results, we took $\xi^*=0.02$, as it offers a fairly good trade-off, reducing considerably the set of maneuvers without a great degradation in terms of energy and accuracy. Table~\ref{tab:maneuvers} shows the final subset $M_r$ of maneuvers, obtained with the selected threshold $\xi^*=0.02$. From now on, we will consider this set $M_r$ of maneuvers in all experiments.

\begin{table}
 \centering
 \begin{tabular}{| c | c | c |} 
 \hline
     $\xi$ & $\delta~(^\circ)$ &  $f~(Hz)$ \\ [0.5ex]
 \hline \hline
0.111 & $-2$ & $0$ \\
 \hline
0.109 & $0$ & $4$ \\
 \hline
0.077 & $0$ & $5$ \\
 \hline
0.076  & $-3$ & $0$ \\
 \hline
0.074 & $-6$ & $0$ \\
 \hline
0.072 & $-5$ & $0$ \\
 \hline
0.057 & $-4$ & $0$ \\
 \hline
0.053 & $-1$ & $0$ \\
 \hline
0.053 & $0$ & $6$ \\
 \hline
0.045 & $-2$ & $6$ \\
 \hline
0.026 & $-3$ & $5$ \\
 \hline
0.026 & $0$ & $0$ \\
 \hline
0.025 & $-4$ & $4$ \\
 \hline
0.023 & $-5$ & $4$ \\
 \hline
0.022 & $-3$ & $4$ \\
 \hline
0.021 & $-6$ & $4$ \\
 \hline
0.021 & $-4$ & $5$ \\
 \hline
\end{tabular}
\caption{Occurrence rate for all maneuvers included in the selected subset $M_r$, obtained with threshold $\xi^*=0.02$.}
\label{tab:maneuvers}
\end{table}

\subsection{Parameters $k_d$ and $k_w$}

After having adjusted the time step and the reference curve for OSPA, we tune parameters $k_d$ and $k_w$ for the pruning operations. We studied the performance of the algorithm varying the values of these two key parameters. For that, we ran OSPA in the 80 scenarios, using the maneuvers in set $M_r$, the reference curve in Equation~\ref{eq:refCurve} and a time step $t_s=12~s$. We analyzed the average values for accuracy $\Delta$, the energy consumption and the computation time.

Table~\ref{tab:tuning_heuristic} depicts the average results of our experiment for some representative values of the parameters. We tested more values within the intervals in Table~\ref{tab:parameters} but having the same trend, so they are not included for the sake of brevity. As expected, the larger the values of $k_d$ and $k_w$, the more node states are explored, and the better the quality of the solutions is, both in terms of energy and accuracy. The computation time also increases though. Depending on the acceptable level of accuracy and the available time budget for OSPA in a given application, different values of $k_d$ and $k_w$ could be selected. 

\begin{table}
 \centering
 \begin{tabular}{| c | c | c | c |} 
 \hline 
 $k_d, k_w$& $\Delta$  &  $E~(W)$  & $Time~(s)$ \\ [0.5ex]
 \hline
 
 \hline
 $10, 15$ & $0.95\pm 0.06$ & $4400\pm 495$ & $137\pm 8$ \\
 
 \hline
 $15, 25$ & $1.53\pm 0.07$ & $4092\pm 424$ & $239\pm 14$ \\
 
 \hline
 $20, 20$ & $0.56\pm 0.08$ & $4270\pm 436$ & $295\pm 18$ \\
 
 \hline
 $25, 35$ & $0.52\pm 0.08$ & $4590\pm 447$ & $443\pm 16$ \\
 \hline
 
\end{tabular}
\caption{Average results and deviations for some representative values of $k_d$ (in meters) and $k_w$. Accuracy, energy and computation time are included.}
\label{tab:tuning_heuristic}
\end{table}

\subsection{Multi-resolution approach}

Finally, we performed another computational experiment of interest to test a multi-resolution approach. OSPA builds a tree generating new states by integrating the system with different maneuvers during a fixed time step $t_s$. However, as we already discussed, that parameter can be key in different aspects. Thus, we also attempted a multi-resolution approach, consisting of having a set with several possible values of the time step instead of a fixed one, i.e., $t_s\in  \mathcal{T}=\{t_1, \dots , t_n\}$. At each leaf node, new nodes are generated for each maneuver using all time steps in $\mathcal{T}$, hence increasing the branching factor and computation time, but also searching space granularity.  

We ran an experiment with the 80 scenarios, using our reference curve in Equation~\ref{eq:refCurve} and the reduced set of maneuvers $M_r$. We set pruning parameters as $k_d=15~m$, $k_w=25$, since they offer a good trade-off in terms of solution quality and computation time according to Table~\ref{tab:tuning_heuristic}. The multi-resolution approach was tested with three sets of time step values:   $\mathcal{T}_1=\{12~s\}$, $\mathcal{T}_2=\{11~s, 13~s\}$, and $\mathcal{T}_3=\{10~s, 12~s, 14~s\}$. Table~\ref{tab:timesteps} shows the average results for the accuracy, the energy consumed and the computation time. Based on the results, we conclude that multiple time step values can improve the energy value with a slight degradation in the accuracy $\Delta$. However, the improvement in solution quality is not that significant in comparison with the outstanding increase in computation time, which made us discard this multi-resolution approach.

\begin{table}
 \centering
 \begin{tabular}{| c | c | c | c |} 
 \hline
 $Steps$ & $\Delta$ & $E~(W)$ & $Time~(s)$\\ [0.5ex]
 \hline
 \hline
 $\mathcal{T}_1$ & $1.53\pm 0.07$ & $4092\pm 424$ & $239\pm 14$ \\
 \hline
 $\mathcal{T}_2$ & $2.01\pm 0.15$ & $3962\pm 439$ & $3150\pm 86$ \\
 \hline
 $\mathcal{T}_3$ & $2.22\pm 0.16$ & $3927\pm 434$ & $7427\pm 243$\\
 \hline
 
\end{tabular}
\caption{Average results and deviations over 80 simulations for the multi-resolution approach with different sets of time steps.}
\label{tab:timesteps}
\end{table}

\section{Experimental Evaluation}
\label{sec:results}

This section shows some experimental results to assess the performance of OSPA. First, we compare OSPA with a probabilistic planner from the literature in order to demonstrate our competitiveness. Then, we depict results of a special case study to illustrate how OSPA can be applied to plan in real time ornithopter trajectories for perching. 

\subsection{Comparison with a sampling approach}
\label{sec:comparison}

We compared the performance of our heuristic-based approach with probabilistic planning approaches, as we believe that these are the most suitable alternatives in the state of the art to tackle online trajectory planning for ornithopters. In particular, we selected the recently published method AO-RRT~\cite{hauser_tro16}, which is, to the best of our knowledge, one of the most competitive in the literature. 
Note that we did not compare it with the purely numerical methods (direct and indirect) for trajectory optimization mentioned in Section~\ref{sec:relatedwork} because their computation times were prohibitive due to the highly nonlinear dynamic model of our ornithopter. 

AO-RRT is an approach to adapt the RRT* probabilistic planner \cite{li2016} to cope with nonlinear dynamics constraints. As the classical RRT*, AO-RRT receives as input an initial state and samples the state space randomly, in order to generate a tree dynamically feasible that eventually will reach the vicinity of the target states. The strategy consists of generating states by integrating numerically the nonlinear system dynamics using control inputs randomly selected. The algorithm stops when a computation time limit is reached and takes as solution a path toward the target state that minimizes a given objective function. In particular, we adapted the original AO-RRT~\footnote{We used the open-source Python implementation provided by the authors.} to use our ornithopter model and the energy consumption as cost function. Moreover, we included a modification to sample control inputs from a discrete set. Otherwise, if we let the algorithm selects frequency values uniformly within the interval $[0,6]~Hz$, the theoretical probability to pick $f=0$ is zero; which would preclude us from performing gliding maneuvers, increasing significantly energy consumption. Therefore, we made the algorithm select control frequencies and tail angles randomly from the discrete set $M$.   

We created a set of 114 simulated scenarios to compare the two approaches. For each simulation, we took $(x_0,z_0)=(0,0)$ as the position of the initial state and selected a target state uniformly within the intervals $x \in [200,250]~m$ and $z \in [-90,20]~m$. Regarding target tolerance, we defined a square of 6-meter side centered in the chosen target point as the region to consider the goal as reached. The parameters selected for OSPA were $t_s=12~s$, $k_d=10~m$, $k_w=15$ (to achieve computation times below $200~s$) and $M_r$ as the set of maneuvers. For AO-RRT, we fixed the maximum computation time to $200~s$; and we checked that OSPA was able to run all the simulated scenarios with average time lower than $200~s$. Another relevant parameter for AO-RRT was the time step between each new control action, as we realized it affected results in several manners. Therefore, we tested different values in our comparison to make it fairer.

\begin{figure}
	\begin{subfigure}[b]{0.5\textwidth}
		\includegraphics[width=7.8cm]{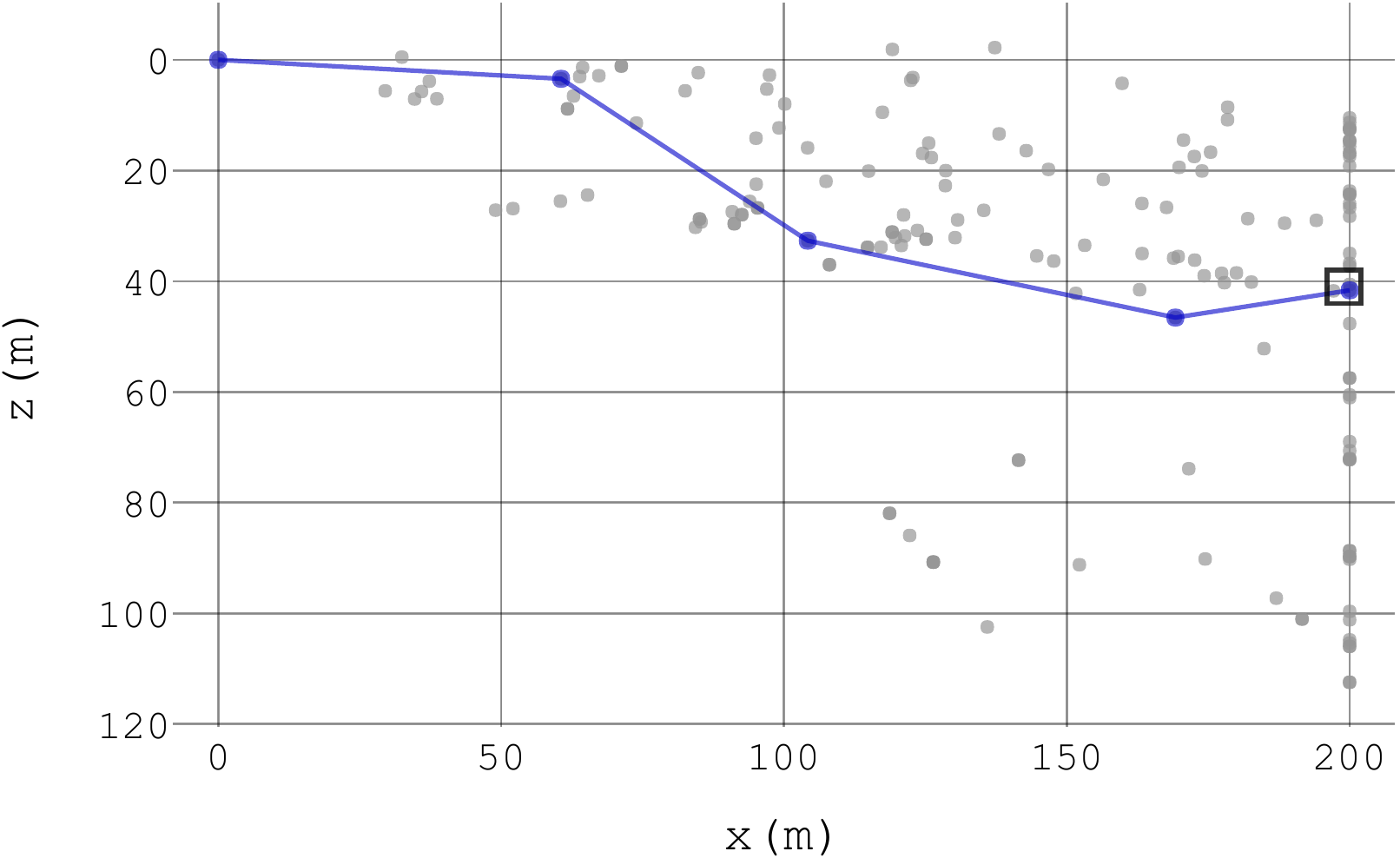}
		\caption{AO-RRT}
		\label{fig:RRTwaypoints} 
	\end{subfigure}
	\quad
	\begin{subfigure}[b]{0.5\textwidth}
		\includegraphics[width=7.8cm]{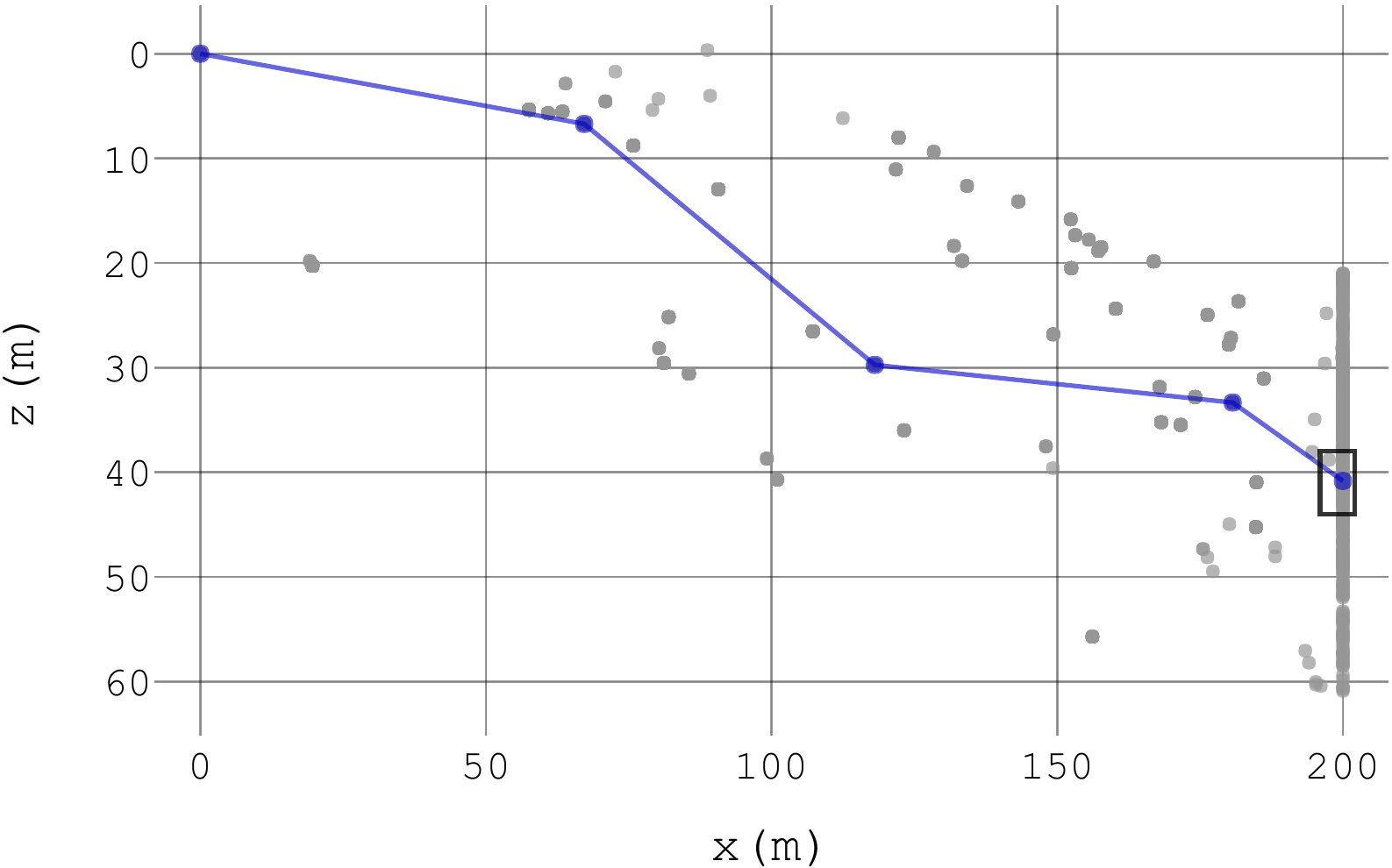}
		\caption{OSPA}
		\label{fig:OSPAwaypoints}
	\end{subfigure}
	\caption{Sampled waypoints in the tree (grey) and solution (blue) for an example experiment. The square denotes the tolerance region around the target state.}
	\label{fig:comparison}
\end{figure}

\begin{figure}
	\centering
	\includegraphics[width=.5\columnwidth]{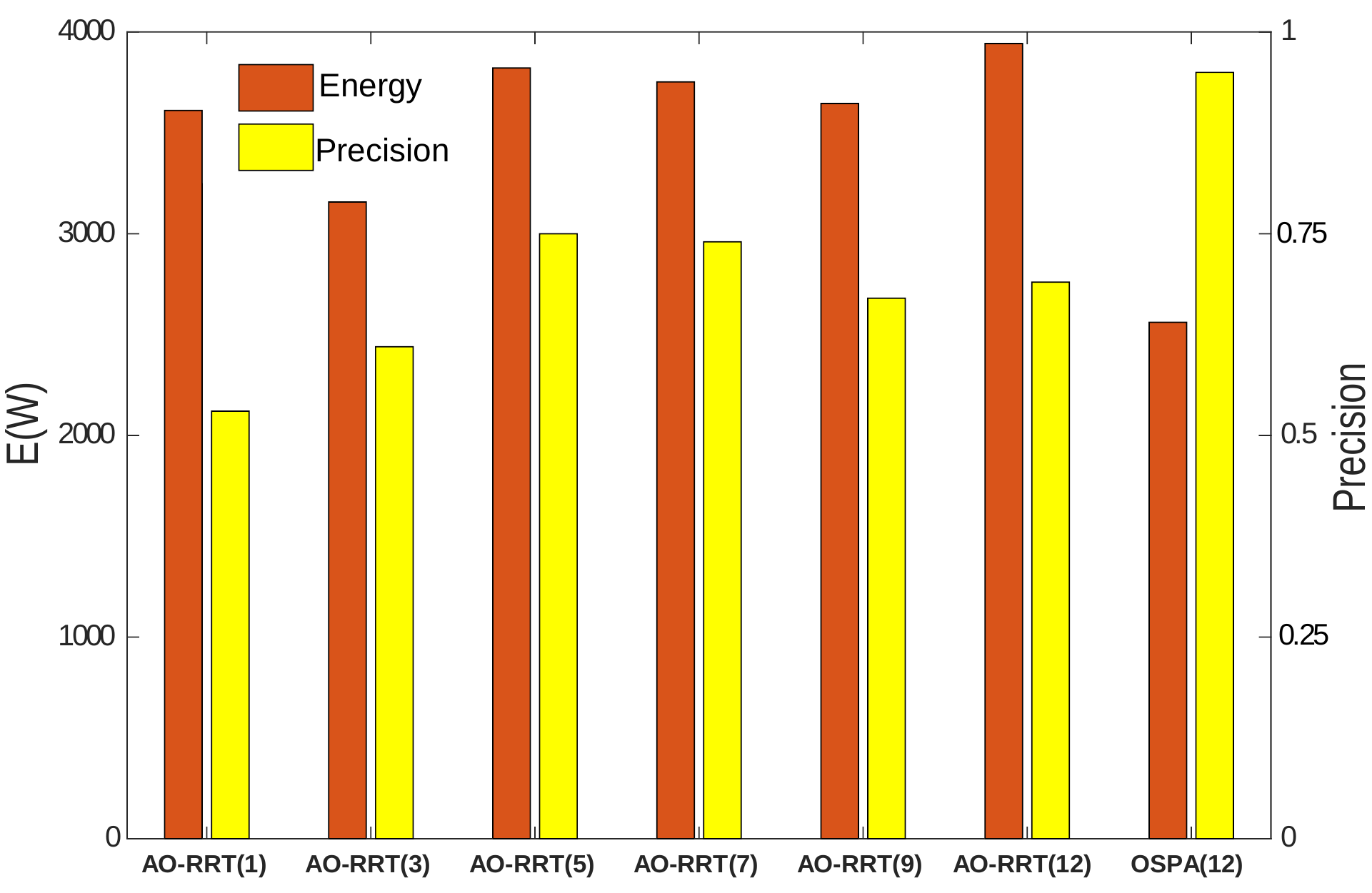}
	\caption{Comparison of energy and precision between OSPA and AO-RRT for different time step values (in parentheses).}
	\label{fig:cmp}
\end{figure}

We used two metrics to compare the approaches: the average energy consumption of the solution trajectory over all scenarios; and the \emph{precision rate}, which is the percentage of cases where a feasible solution was found within the 6-squared-meter tolerance region.
Figure~\ref{fig:comparison} shows the output of one of the simulated experiments for AO-RRT and OSPA, including the obtained solutions and the sampled waypoints in the trees. The figure depicts the advantages of the search strategy in OSPA; AO-RRT samples random states more uniformly distributed in the space, while in OSPA, thanks to the reference curve, states are more concentrated around the optimal solution.    
Figure~\ref{fig:cmp} depicts the resulting metrics for the complete comparison. It can be seen that the precision rate of OSPA is always greater. Also, OSPA achieves more efficient trajectories in terms of energy. Thus, OSPA outperforms the state-of-the-art AO-RRT algorithm for ornithopter trajectory optimization.   

\subsection{A case study: landing for perching}
\label{sec:caseStudy}

OSPA is thought for ornithopter navigation through short and medium distance flights, and it aims to optimize energy consumption during operation. For those scenarios, we can compute a trajectory online and then use nonlinear controllers at high rate for trajectory following. Now, we show a particular case study to demonstrate how OSPA can also be used for landing maneuvers when the ornithopter is going to perch. 

A perching maneuver was studied analytically in \cite{crowther2000perched}, consisting of two phases: a gliding phase (almost horizontal in this case) and a rapid pitch up to a high angle of attack. The example in Figure~\ref{fig:segmentation} shows various stages of this type of maneuvers. Perching maneuvers entail computing trajectories quite fast, as the ornithopter has no much time for reaction. Also, it is more relevant to land closer to the target rather than reducing the energy cost, as perching requires of high accuracy, particularly if the available area for landing is small. Therefore, we measure in this experiment the Euclidean distance to the target state considering only position, as vehicle velocities and attitude can be sensitive and may be adjusted by lower-level attitude controllers at high rate. 

We created a set of experiments to test OSPA computing trajectories to a landing spot. Typically, the initial and target positions should be relatively close and the altitude be descendant. Therefore, we set the ornithopter initially at the origin of coordinates with zero pitch and the target point at a 10-meter longitudinal distance with $z$ coordinate ranging between $2$ and $5~m$. Recall that positive $z$ values mean descending.  

Since we aim at lower computation time when planning perching trajectories, we refine the set of maneuvers to apply OSPA. We experimentally selected a new reduced set of maneuvers for perching $M_p$, focusing specially on those with particular interest for that operation. To do this, we ran our set of experiments and selected $M_p$ following the same procedure as in Section~\ref{section:maneuvers}. In particular, we used a probability threshold $\xi^*=0.03$ and obtained the reduced set $M_p$ shown in Table ~\ref{tab:perchingmaneuvers}. It can be seen that most of these maneuvers involve gliding as it was demonstrated in~\cite{cory2008experiments} for optimized perching maneuvers. 

\begin{table}
 \centering
 \begin{tabular}{| c | c | c |} 
 \hline
     $\xi$ & $\delta~(^\circ)$ &  $f~(Hz)$ \\ [0.5ex]
 \hline\hline
0.079 & $-1$ & $0$ \\
 \hline
0.107 & $-2$ & $0$ \\
 \hline
0.122 & $-3$ & $0$ \\
 \hline
0.081  & $-4$ & $0$ \\
 \hline
0.090 & $-5$ & $0$ \\
 \hline
0.093 & $-6$ & $0$ \\
 \hline
0.115 & $0$ & $4$ \\
 \hline
0.057 & $0$ & $5$ \\
 \hline
0.033 & $0$ & $6$ \\
 \hline
\end{tabular}
\caption{Maneuvers used for perching planning. This set $M_p$ was obtained with a threshold $\xi^*=0.03$.}
\label{tab:perchingmaneuvers}
\end{table}

We set OSPA parameters to $t_s=1~s$, $k_d=2~m$ and $k_w=4$.
Table~\ref{tab:preperching} shows results for the experiments performed. It is important to remark that, in all cases, the planning time is around $1~s$ and the distance error with respect to the target below $0.05~m$. Interestingly, the optimized strategy yielded by OSPA was similar to the profile showed in Figure~\ref{fig:segmentation}, i.e., a
gliding phase followed by a pitch up with maximum upward elevator deflection. This indicates that OSPA can be a valid approach in practice for computing trajectories to approach a landing area where the ornithopter decides to perch.

\begin{table}
 \centering
 \begin{tabular}{| c | c | c |} 
 \hline
     $z_f~(m)$ & Error (m) &  Time (s) \\ [0.5ex]
 \hline\hline
$2$ & $0.049$ & $1.09$ \\
 \hline
$2.5$ & $0.005$ & $1.08$ \\
 \hline
$3$ & $0.021$ & $0.98$  \\
 \hline
$3.5$ & $0.008$ & $1.10$  \\
 \hline
$4$  & $0.042$ & $1.08$ \\
 \hline
$4.5$  & $0.017$ & $1.11$ \\
 \hline
$5$  & $0.033$ & $0.95$ \\
 \hline
\end{tabular}
\caption{Results for perching experiments using OSPA. The final point is located 10 meters away in longitudinal distance, and at different altitudes $z_f$. Distance error to the target position and computational time are shown.}
\label{tab:preperching}
\end{table}

\section{Discussion and Future Work}
\label{sec:conclusions}

This work proposed OSPA, a new algorithm for kinodynamic planning of trajectories for autonomous ornithopters. The method is able to compute energy-efficient trajectories in an online fashion, combining gliding and flapping maneuvers. 
OSPA builds trees dynamically feasible and runs heuristic search to plan trajectories. This paradigm can be applied to any dynamic model (we used a nonlinear aerodynamic model for ornithopters) and different flight types. We demonstrated a proper performance of the algorithm for medium distance flights of up to $250~m$, but also for planning short landing trajectories of up to $10~m$. Computation time is suitable for online trajectory planning, achieving solutions for short flights in less than $1~s$. Moreover, our experimental results showed that OSPA outperforms alternative probabilistic kinodynamic planners both in cost (total energy) and accuracy (distance to the target). An open-source implementation of OSPA and our benchmarks is available online~\footnote{\url{https://github.com/fragnarxx/kinodynamic-planning}}. Notice that our current implementation is written in Python, so there is still room for improvement with more efficient languages like C. 

As future work, we mention some potential extensions for OSPA that we discuss in the following.

{\it Improving heuristic performance:} One of the main aspects that affects the performance of the heuristic search in OSPA is the reference curve, as the algorithm relies on having a good approximation of optimal curves to guide search. We proposed curves computed empirically, but an open mathematical problem is to calculate the theoretical reference curve that minimizes the energy consumption for a given model. With a better estimation of that optimal energy curve, the parameters $k_d$ and $k_w$ could be reduced to compute pseudo-optimal trajectories more efficiently.

{\it Planning in dynamic scenarios:}
OSPA can be used for trajectory planning in dynamic scenarios, recomputing trajectories online as the environment changes. Nonetheless, another problem to explore would be to address scenarios with static or mobile obstacles, not only open spaces. This could be done by combining state-of-the-art collision avoidance algorithms with OSPA, or integrating some procedure for collision checking within the planner.

{\it Planning in 3D:}
In this paper, we considered 2D trajectories for the ornithopter. However, OSPA is not limited to that, as ornithopter models in 3D could be also used with OSPA to plan 3D trajectories. This is particularly useful for settings where wind effects on lateral displacement cannot be neglected, and it could be tackled by considering curvature-constrained trajectories as in~\cite{al2013wind,zhang2014memetic}.

{\it Machine learning methods:}
These days methods for trajectory planning based on machine learning are spreading fast. One of the issues is the lack of data of real bird flights in order to imitate their trajectories with ornithopters. In this sense, OSPA may be helpful to generate artificial bio-inspired datasets with pseudo-optimal trajectories, which could be used for training alternative machine learning methods. 

{\it Experiments with real ornithopter:}
Finally, we plan to use OSPA to compute trajectories on board our real ornithopter prototype within the framework of the GRIFFIN project. For that, we will combine the planner with nonlinear controllers for flight stabilization and trajectory following. 

\vspace{.5cm}


\bibliographystyle{plain}
\bibliography{main-arxiv}
\end{document}